\pdfoutput=1

\documentclass[11pt]{article}

\usepackage[final]{acl}

\usepackage{times}
\usepackage{latexsym}

\usepackage[T1]{fontenc}

\usepackage[utf8]{inputenc}

\usepackage{microtype}

\usepackage{inconsolata}

\usepackage{graphicx}
\usepackage{inconsolata}
\usepackage{amsmath}
\usepackage{multirow}
\usepackage{xfrac}
\usepackage{xspace}
\usepackage{comment}

\newcommand{\our}{\text{MixORE}\xspace}

\title{Towards a More Generalized Approach in Open Relation Extraction}

\author{Qing Wang, Yuepei Li, Qiao Qiao, Kang Zhou, Qi Li\\
        Department of Computer Science, Iowa State University, Ames, Iowa, USA\\ 
        \texttt{\{qingwang, liyp0095, qqiao1, kangzhou, qli\}@iastate.edu}}
        
\begin{document}
\maketitle
\begin{abstract}
Open Relation Extraction (OpenRE) seeks to identify and extract novel relational facts between named entities from unlabeled data without pre-defined relation schemas. Traditional OpenRE methods typically assume that the unlabeled data consists solely of novel relations or is pre-divided into known and novel instances. However, in real-world scenarios, novel relations are arbitrarily distributed. In this paper, we propose a generalized OpenRE setting that considers unlabeled data as a mixture of both known and novel instances. To address this, we propose MixORE, a two-phase framework that integrates relation classification and clustering to jointly learn known and novel relations. Experiments on three benchmark datasets demonstrate that MixORE consistently outperforms competitive baselines in known relation classification and novel relation clustering. Our findings contribute to the advancement of generalized OpenRE research and real-world applications. Source code is available\footnote{\url{https://github.com/qingwang-isu/MixORE}}.
\end{abstract}

\section{Introduction} \label{sec:intro}


Open Relation Extraction (OpenRE) is a fundamental task in Information Extraction (IE) that aims to identify and extract relational facts between named entities from unlabeled data. Unlike traditional Relation Extraction (RE), which relies on a predefined set of relations and requires end-users to specify their information needs and provide costly annotations, OpenRE operates in a more flexible ``open-world'' setting. It proactively discovers novel relations, generalizes them into meaningful categories, and identifies additional instances, making it a more adaptable approach for large-scale IE.


In recent years, OpenRE has attracted increasing attention from researchers. \citet{DBLP:conf/emnlp/WangZLLZW22} and \citet{DBLP:conf/emnlp/LiJH22} introduce prompt-based learning methods and advanced clustering techniques, achieving impressive results on unlabeled data. However, existing OpenRE methods typically assume either that the unlabeled data consists entirely of novel relations or that there is prior information indicating whether an instance belongs to a known or novel relation. These assumptions do not accurately reflect the complexities of real-world scenarios. 

\citet{DBLP:conf/emnlp/HoganLS23} further dispose of the simplifying assumptions and make new assumptions that the unlabeled data includes known and novel instances and that novel relations are typically rare, belong to the long-tail distribution, and tend to be explicitly expressed. Their model, KNoRD, is built around these assumptions. However, the ``long-tail'' assumption may not always hold, particularly in scenarios where novel relations emerge as newly-recognized concepts in the real world that have not yet been labeled. Additionally, novel relations may arise when human annotators label only some relations within a large dataset, leaving many potential relations unidentified. For novel relations that do not follow the long-tail distribution, KNoRD tends to introduce additional noise and its performance degrades.
Furthermore, we observe that a noticeable performance gap still exists between known and novel instances \citep{DBLP:conf/emnlp/HoganLS23}, highlighting the potential for further OpenRE research.

In this paper, we relax the ``long-tail'' assumption and instead assume the unlabeled data contains both known and novel instances, with no restrictions on the nature of these relations. We propose \our model to effectively classify known instances and identify novel relations within unlabeled data. \our has two phases: novel relation detection and open-world semi-supervised joint learning (OW-SS joint learning).
 
In the first phase, our goal is to identify potential novel relations within unlabeled data.
We represent each known relation with a one-hot vector in latent space and train a Semantic Autoencoder (SAE) \citep{DBLP:conf/cvpr/KodirovXG17} on labeled data. The trained SAE then maps both labeled and unlabeled instances into the shared latent space, where known instances will cluster around their respective one-hot vectors. In contrast, novel instances, which are less likely to align with any known relations, tend to appear as outliers in this mapping process. Furthermore, instances in the same novel relation often exhibit a clustering pattern.
Therefore, we leverage each unlabeled instance's similarity to the known relation one-hot vectors as a criterion for outlier detection. Subsequently, we apply the Gaussian Mixture Model (GMM) \citep{DBLP:journals/jmlr/PedregosaVGMTGBPWDVPCBPD11} to cluster these outliers into novel relation groups and extract instances closest to each cluster centroid as high-quality weak labels for further training.

In the second phase, OW-SS joint learning, we utilize weak labels and adopt a continual learning strategy to align our approach with the evolving nature of OpenRE in real-world applications. \our is designed based on the insight that classifying known relations requires learning compact and well-separated feature representations, whereas detecting novel relations benefits from capturing diverse and transferable features. To achieve this, we incorporate contrastive learning by leveraging both labeled instances and data distribution to form positive pairs and propose the OW-SS loss function, which jointly optimizes relation classification and clustering.

In summary, our main contributions are:
\begin{itemize}
\item We comprehensively review the assumptions made in previous OpenRE studies and introduce a generalized OpenRE setting. 
\item We propose a two-phase framework \our that learns discriminative features for known relations while continuously incorporating novel information from unlabeled data, making it more adaptable to OpenRE in real-world scenarios.
\item Experimental results demonstrate that our approach achieves remarkable performance on both known and novel relations across the FewRel, TACRED, and Re-TACRED datasets.
\end{itemize}

\section{Related Work}
Relation Extraction (RE) is an essential Natural Language Processing (NLP) task and has been extensively studied with approaches relying on supervised learning techniques trained on manually annotated datasets \citep{DBLP:conf/muc/MillerCFRSSW98,DBLP:conf/emnlp/ZelenkoAR02,DBLP:journals/tacl/PengPQTY17,DBLP:conf/naacl/ZhongC21,DBLP:conf/acl/WadhwaAW23}. While RE models achieve high performance, their dependency on large-scale labeled data presents a major limitation \citep{DBLP:conf/aaai/ZhouQL023}.
Moreover, they operate under a ``closed-world'' assumption, where relations are pre-defined, limiting their ability to handle emerging or novel relations. To address these challenges, OpenRE is proposed to proactively identify novel relations from unlabeled data in an ``open-world'' setting, making it more suitable for real-world, large-scale information extraction. 

Existing OpenRE methods mostly operate under two settings. The first setting is unsupervised relation extraction (URE), where models identify relations between named entities from unlabeled data without relying on manual annotations. \citet{DBLP:conf/naacl/LiuHZLWY22} propose a hierarchical exemplar contrastive learning framework that refines relation representations by leveraging both instance-level and exemplar-level signals for optimization. \citet{DBLP:conf/emnlp/WangZQLL23} strengthen the discriminative power of contrastive learning with both within-sentence pairs augmentation and augmentation through cross-sentence pairs extraction to increase the diversity of positive pairs.

The second setting of existing methods, semi-supervised OpenRE, involves training models on labeled data with known relations, while the unlabeled data consists entirely of novel relations or is pre-divided into known and novel sets. \citet{DBLP:conf/emnlp/WangZLLZW22} develop a novel prompt-based framework that enables the model to generate efﬁcient representations for instances in the open
domain and learn clustering novel relational instances. \citet{DBLP:conf/emnlp/LiJH22} design a co-training framework that combines the advantage of type abstraction and
the conventional token-based representation. 

\begin{figure*}
\centering
\includegraphics[width=6.4in]{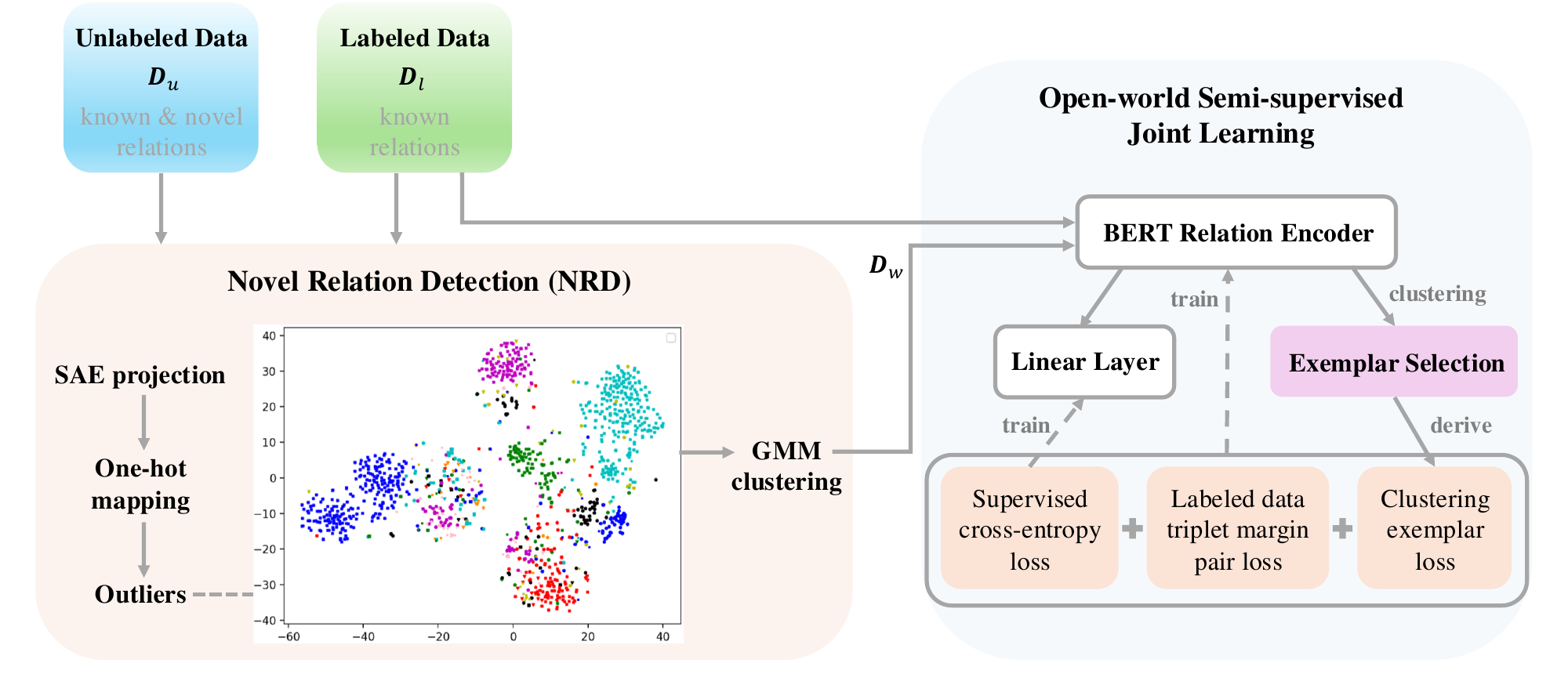}
\caption{Overview of \our Framework.}
\label{fig:overview}
\end{figure*}

There are some recent studies trying to address the open-world semi-supervised learning (Open-world SSL) setting, where unlabeled data contains a mixture of both known and novel classes.
\citet{DBLP:conf/iclr/CaoBL22} propose ORCA for computer vision tasks. This method introduces uncertainty adaptive margin loss objective to either classify unlabeled image instances into one of the known classes or discover novel classes and assign instances to them. \citet{DBLP:conf/emnlp/HoganLS23} later introduce KNoRD for open-world relation extraction.
With prompt-based training, KNoRD effectively classifies explicitly and implicitly expressed relations from known and novel relations within unlabeled data. However, the authors assume novel relations are typically rare and belong to the long-tail distribution. In this study, we relax this assumption and instead assume the unlabeled data comprises both known and novel instances, where the known and novel relations can be arbitrary.

\section{Task Formulation}
We formalize the OpenRE task as follows. Let $\boldsymbol{x}=[x_1,...,x_n]$ denote a sentence, where $x_i$ represents the $i$-th $(1\leq i\leq n)$ token. 
In the sentence, a named entity pair $(\boldsymbol{e}_h, \boldsymbol{e}_t)$ is recognized in advance, where $\boldsymbol{e}_h$ represents the head entity and $\boldsymbol{e}_t$ represents the tail entity.
Let $\boldsymbol{\mathcal{D}}_l=[(\boldsymbol{x}^1, \boldsymbol{y}^1),...,(\boldsymbol{x}^M,\boldsymbol{y}^M)]$ be the labeled data consists of $M$ instances with the corresponding sentences, the target entity pairs, and relation labels. Let $\boldsymbol{\mathcal{D}}_u=[(\boldsymbol{x}^1),...,(\boldsymbol{x}^N)]$ be the unlabeled data consists of $N$ instances with only corresponding sentences and target entity pairs. We denote the set of relations in the labeled data as $\boldsymbol{\mathcal{C}}_{known}$ and the set of relations in the unlabeled test data as $\boldsymbol{\mathcal{C}}_u$. Following \citet{DBLP:conf/iclr/CaoBL22}, we assume category/class shift $\boldsymbol{\mathcal{C}}_{known}\subseteq\boldsymbol{\mathcal{C}}_u$ (i.e., the relations encountered at test time may not have been explicitly labeled or seen during training). We define the set of novel relations $\boldsymbol{\mathcal{C}}_{novel}=\boldsymbol{\mathcal{C}}_u-\boldsymbol{\mathcal{C}}_{known}$.

The goal of OpenRE is to assign known instances in $\boldsymbol{\mathcal{D}}_u$ to their respective known relations $\boldsymbol{\mathcal{C}}_{known}$, while also identifying $|\boldsymbol{\mathcal{C}}_{novel}|$ novel relation clusters, where
$|\boldsymbol{\mathcal{C}}_{novel}|$ represents the number of novel relations in the corpus.

\section{Methodology}

In this section, we introduce the proposed \our, a two-phase framework that integrates relation classification and clustering to jointly learn known and novel relations. Our methodology incorporates novel relation detection for obtaining weak labels and open-world semi-supervised joint learning (OW-SS joint learning) to progressively refine the model. Figure \ref{fig:overview} provides an overview of the framework.

\subsection{Relation Encoder} \label{sec:rel}
Given a sentence along with its named entities and entity types, the relation encoder generates a vector representation that captures the relationship between the entities. To highlight the entities of interest, we adopt entity marker tokens, a widely used technique in relation extraction models \citep{DBLP:conf/acl/SoaresFLK19,DBLP:conf/emnlp/XiaoYXHLSLL20,DBLP:conf/naacl/LiuHZLWY22,DBLP:conf/emnlp/WangZQLL23}.

Specifically, for a given sentence $\boldsymbol{x}=[x_1,...,\boldsymbol{e}_h,...,\boldsymbol{e}_t,...,x_n]$,
we insert <$e1$:type> and </$e1$:type> to denote the beginning and end of the head entity $\boldsymbol{e}_h$, and similarly, <$e2$:type> and </$e2$:type> for the tail entity $\boldsymbol{e}_t$, where "type" is replaced with the actual entity type. We use BERT$_{base}$ model \citep{DBLP:conf/naacl/DevlinCLT19} to obtain the contextualized sentence representation $\boldsymbol{h}$. To effectively capture relational context and enhance focus on the target entity pair, we derive the following fixed-length relation representation:
\begin{equation}
\boldsymbol{h}_r=[h_{\textit{<}e1\textit{:type>}}|h_{\textit{<}e2\textit{:type>}}] 
\end{equation}
to express the relation between the marked entities in $\boldsymbol{x}$ , where $|$ denotes the concatenation.

\subsection{Novel Relation Detection} \label{sec:nrd}
In the first phase of \our, our objective is to identify potential novel relations within unlabeled data. These novel relations, once detected, can be leveraged as weak labels to enhance the training process, particularly in real-world scenarios where labeled data is scarce or unavailable. 


Our novel relation detection approach is founded on the assumption that known instances naturally cluster around their respective relation centroids, forming well-defined groups. In contrast, novel instances, which do not correspond to any known relations, are likely to appear as outliers.
However, in practice, the lack of labeled data for novel relations results in ambiguous feature representations, making it challenging to differentiate between known and novel relations.
Additionally, clustering algorithms such as K-Means and Gaussian Mixture Models (GMM) often struggle with high-dimensional feature spaces, further complicating the task of accurately grouping novel instances.

\begin{figure}
\centering
\includegraphics[width=0.40\textwidth]{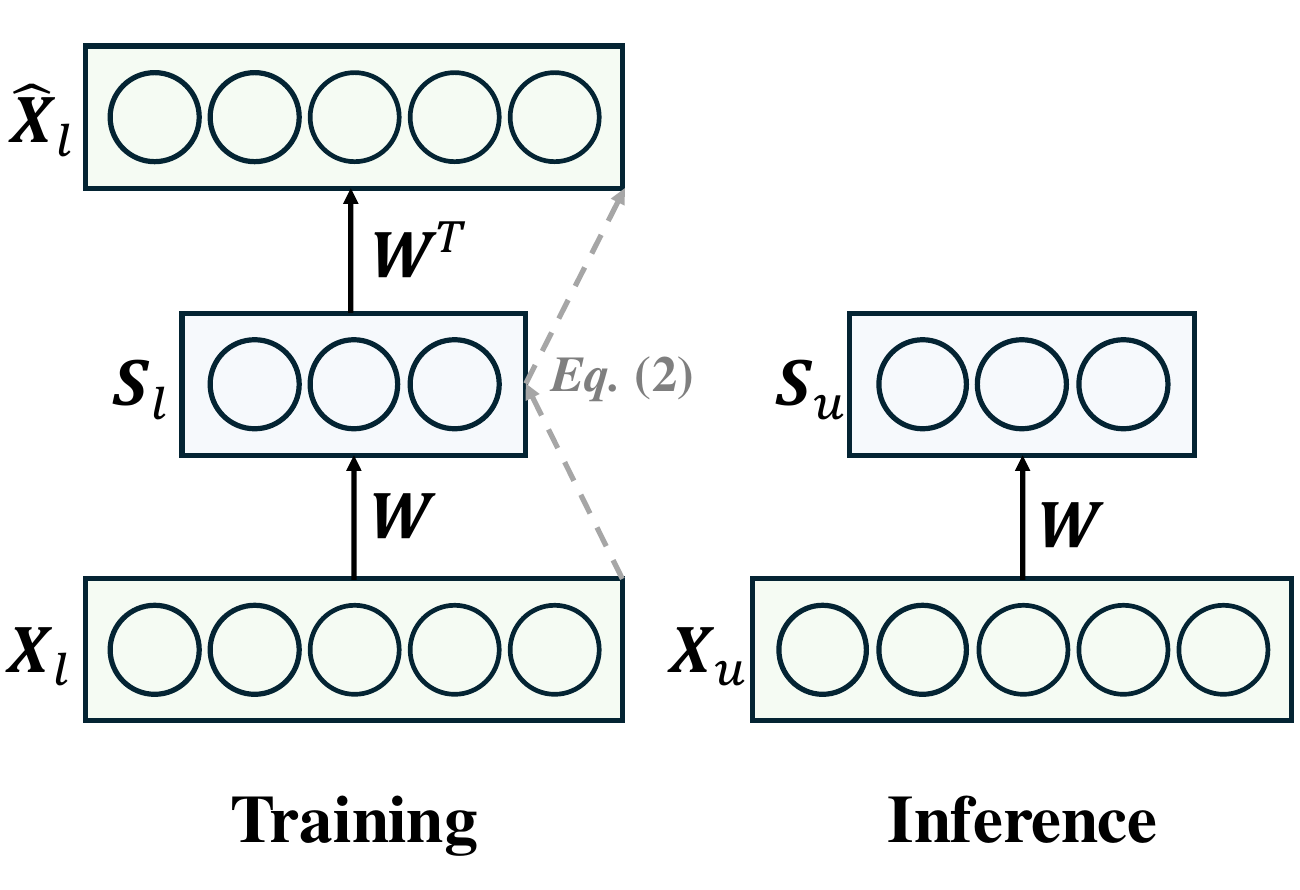}
\caption{Semantic Autoencoder (SAE)}
\label{fig:SAE}
\end{figure}

To effectively learn a low-dimensional projection function that generalizes well to both known and novel relations, we employ the encoder-decoder paradigm. In this approach, the encoder maps a feature vector into an intermediate low-dimensional space, while the decoder imposes an additional constraint by ensuring that the projected representation can accurately reconstruct the original feature vector. Specifically, we adopt the Semantic Autoencoder (SAE) \citep{DBLP:conf/cvpr/KodirovXG17}, a simple and extremely efficient architecture, as illustrated in Figure \ref{fig:SAE}.

The labeled data $\boldsymbol{\mathcal{D}}_l$ is first processed by the relation encoder (defined in Sec. \ref{sec:rel}), generating the input data matrix $\boldsymbol{X}_l=[(\boldsymbol{h}_r^1),...,(\boldsymbol{h}_r^M)]$. It is projected into a latent space of $|\boldsymbol{\mathcal{C}}_{known}|$ dimensional with a projection matrix $\boldsymbol{W}$. 
The latent space is constrained to serve as a semantic representation space. 
To enforce independence among relations, we incorporate one-hot vectors to encode known relations and obtain the latent representation $\boldsymbol{S}_l$. To further simplify the model, we use tied weights, that is, the transposed projection matrix $\boldsymbol{W}^T$ projects the latent representation $\boldsymbol{S}_l$ back to the feature space, and becomes $\hat{\boldsymbol{X}_l}$. The learning objective is as follows:
\begin{align} \label{eq:obj}
    \min_{\mathbf{W}} \ & \| \mathbf{X}_l - \mathbf{W}^\top \mathbf{S}_l \|_F^2 + \lambda\| \mathbf{W} \mathbf{X}_l- \mathbf{S}_l \|_F^2,
\end{align}
where $\lambda$ is a weighting coefficient that balances the contributions of the first and second terms, corresponding to the losses of the decoder and encoder, respectively. Following \citet{DBLP:conf/cvpr/KodirovXG17}, we efficiently derive the optimal solution for $\boldsymbol{W}$ with Bartels-Stewart algorithm \citep{DBLP:journals/cacm/BartelsS72}, a closed-form solver that eliminates iterative updates and thus accelerates computation. To keep the first phase lightweight, we leave the BERT$_{base}$ parameters in the relation encoder frozen.


During inference, we input all unlabeled data $\boldsymbol{\mathcal{D}}_u$ into the relation encoder, and subsequently pass the resulting relation representations $\boldsymbol{X}_u$ through the encoder of the SAE to obtain the latent representation $\boldsymbol{S}_u$. For each vector $v$ in $\boldsymbol{S}_u$, we calculate its cosine similarity with each known relation one-hot vector and record the highest similarity score as its mapping score. This process assigns each instance in $\boldsymbol{\mathcal{D}}_u$ to the most probable known relation. In line with the conventional 5\% significance level used in statistical hypothesis testing, we designate the 5\% of unlabeled instances with the lowest mapping scores as outliers. 

Instances belonging to the same novel relation also tend to cluster together. We subsequently employ the Gaussian Mixture Model (GMM) \citep{DBLP:journals/jmlr/PedregosaVGMTGBPWDVPCBPD11} to cluster these outliers into $|\boldsymbol{\mathcal{C}}_{novel}|$ novel relation clusters. GMM assumes that the data points are generated from a mixture of several Gaussian distributions, each representing a cluster. The model defines the probability density function (PDF) of the data as:
\begin{align}
    p(v|\Theta) & = \sum_{i=1}^{|\boldsymbol{\mathcal{C}}_{novel}|}\pi_i\mathcal{N}(v|\mu_i,\Sigma_i),
\end{align}
where $p(v|\Theta)$ is the likelihood of observing data point $v$, $\pi_i$ is the mixture weight of the $i$-th Gaussian component, and $\mathcal{N}(v|\mu_i,\Sigma_i)$ represents the multivariate Gaussian distribution with mean $\mu_i$ and covariance matrix $\Sigma_i$. 

To extract high-quality weak labels for subsequent training, we select instances closest to each cluster centroid. Specifically, we retain instances with a GMM posterior probability greater than 0.95, ensuring that only those with high confidence in their cluster assignments are used as weak labels.
The resulting set of weakly-labeled instances is denoted as $\boldsymbol{\mathcal{D}}_w$.

\subsection{Open-world Semi-supervised Joint Learning}

The ultimate goal of the Open-world SSL setting is to adaptively expand the model's understanding of novel relations while preserving high performance on known relations. To effectively handle both known and novel relations, we employ a continual learning \citep{DBLP:journals/pami/WangZSZ24} strategy. In the OW-SS joint learning phase, the proposed \our model is first warmed up by training on $\boldsymbol{\mathcal{D}}_l$, which consists of known relations only. Following the rehearsal-based strategy in Continual Relation Extraction \citep{DBLP:conf/acl/CuiYYHCYX20,DBLP:conf/acl/WuLZGZ024}, \our is continually trained on both labeled known instances $\boldsymbol{\mathcal{D}}_l$ and the weakly-labeled novel instances $\boldsymbol{\mathcal{D}}_w$. 

First, for relation classification, we employ the following cross-entropy loss function: 
\begin{align} \label{eq:lc}
    \mathcal{L}_{c} & = -\frac{1}{D_c}\sum_{i=1}^{D_c}\sum_{r=1}^{|\boldsymbol{\mathcal{C}}_u|} y_r^ilog(\hat{y_r^i}),
\end{align}
where $y_r^i$ is 1 if sample $i$ belongs to relation $r$ otherwise 0, $\hat{y_r^i}$ is the predicted probability of sample $i$ belongs to relation $r$, $|\boldsymbol{\mathcal{C}}_u|=|\boldsymbol{\mathcal{C}}_{known}|+|\boldsymbol{\mathcal{C}}_{novel}|$ is the number of relations in the unlabeled data, and $D_c$ represents the number of labeled instances in the current epoch.

\citet{DBLP:conf/nips/ZhangJFWZWT0G22} demonstrate that, in computer vision tasks, discriminative features are preferred for classifying known classes, whereas rich and diverse features are essential for identifying novel classes. Such findings should also apply to OpenRE tasks, as classifying known relations requires learning compact and well-separated feature representations, while detecting novel relations benefits from capturing diverse and transferable features that generalize beyond the labeled data. 

Contrastive Learning, a strategy widely adopted by state-of-the-art RE models \citep{DBLP:conf/naacl/LiuHZLWY22,DBLP:conf/emnlp/WangZQLL23,DBLP:conf/acl/WuLZGZ024}, enhances relation representations by pulling semantically similar relation sentences (positive pairs) closer while pushing apart sentences with different relations (negative pairs).
We integrate contrastive learning using two strategies to form positive pairs: sampling from labeled instances and leveraging the data distribution. This approach enables us to jointly capture classification signals and the underlying data distribution, leading to more robust relation representations.

We begin by utilizing labeled data to construct positive pairs. Since weak labels can be noisy, to minimize the risk of introducing false positive pairs from $\boldsymbol{\mathcal{D}}_w$, we restrict the generation of positive pairs to $\boldsymbol{\mathcal{D}}_l$. Specifically, we sample instances from $\boldsymbol{\mathcal{D}}_l$ such that two instances sharing the same relation form a positive pair and ensure that each relation has an equal number of positive pairs sampled, except in cases where there are insufficient instances to enumerate. Let $\boldsymbol{\mathcal{P}}=[(\boldsymbol{a_r}^1,\boldsymbol{p_r}^1),...,(\boldsymbol{a_r}^{D_m},\boldsymbol{p_r}^{D_m})]$ denote the set of relation representations of the sampled $D_m$ positive pairs. In this work, we fix the number of sampled positive pairs to $D_m=5D_c$. As noted by \citet{DBLP:conf/emnlp/WangZQLL23}, relation semantics between two sentences should not be treated as a strict ``same/different'' distinction but rather as a similarity spectrum. To handle this, the triplet margin loss function for the labeled data positive pairs is defined as:
\begin{equation}
\begin{aligned}
\mathcal{L}_{lm} & = \frac{1}{D_m}\sum_{i=1}^{D_m} \max\{dist(\boldsymbol{a}_r^i, \boldsymbol{p}_r^i)\\ &-dist(\boldsymbol{a}_r^i, \boldsymbol{n}_r^i)+\gamma, 0\},
\end{aligned}
\end{equation}
where $dist(\cdot)$ denotes the cosine distance function, $\boldsymbol{n}_r^i$ represents a randomly sampled negative example for $\boldsymbol{a}_r^i$, and $\gamma$, known as the margin, is a hyperparameter.

To further incorporate data distribution, we encourage relation representations to align more closely with their respective cluster centroids while pushing them away from other clusters. Each instance and its corresponding virtual centroid are treated as a positive pair, reinforcing the cluster structure in the representation space. Following \citet{DBLP:conf/naacl/LiuHZLWY22}, we select relational exemplars at multiple granularities by computing cluster centroids for different values of $k$ using K-Means algorithm. These exemplars dynamically adjust in response to parameter updates in the relation encoder during each training epoch. Since an instance either belongs to a cluster or not, we use the following clustering exemplar loss function:
\begin{equation}
\mathcal{L}_{e} = -\sum_{i=1}^{D_c} \frac{1}{L} \sum_{l=1}^L log \frac{exp(\boldsymbol{h}_r^i \cdot \boldsymbol{e}_j^l/\tau)}{\sum_{q=1}^{c_l}exp(\boldsymbol{h}_r^i \cdot \boldsymbol{e}_q^l/\tau)},
\end{equation}
where $j\in[1, c_l]$ represents the $j$-th cluster at granularity layer $l$, $\boldsymbol{e}_j^l$ is relation representation of the exemplar of instance $i$ at layer $l$, and $\tau$ is a is a temperature hyperparameter \citep{DBLP:conf/cvpr/WuXYL18}.

Our overall OW-SS loss function is defined as the addition of classification loss $\mathcal{L}_{c}$, labeled data triplet margin loss $\mathcal{L}_{lm}$, and clustering exemplar loss $\mathcal{L}_{e}$:
\begin{equation} \label{eq:total}
\mathcal{L} = \mathcal{L}_{c} + \mathcal{L}_{lm} + \mathcal{L}_{e},
\end{equation}
which jointly optimizes relation classification and clustering. The model architecture consists of a BERT encoder followed by a fully connected linear layer. The BERT encoder is fine-tuned using Eq. \ref{eq:total}, while the linear layer parameters are updated based on Eq. \ref{eq:lc}.

\subsection{Inference}
During inference, each instance is encoded using the trained model to obtain its relation representation and predicted label. If the predicted label corresponds to a known relation, it is directly accepted as the final result. However, since novel relations are trained with weak labels, their predicted labels may not be accurate.
Therefore, we leverage relation representations for novel relations instead. Specifically, we employ Faiss K-Means clustering algorithm \cite{DBLP:journals/tbd/JohnsonDJ21}, an efficient implementation of K-Means optimized for large-scale and high-dimensional data, to cluster these relation representations and assign relations based on the clustering results.

\subsection{Data Augmentation}
Several studies have demonstrated that data augmentation can significantly enhance the performance of RE models \citep{DBLP:conf/acl/LiuYLH020,DBLP:conf/naacl/LiuHZLWY22,DBLP:conf/emnlp/WangZQLL23}. In this work, we apply the data augmentation technique proposed by \citet{DBLP:conf/emnlp/WangZQLL23}, which leverages within-sentence pairs augmentation and augmentation through cross-sentence pairs extraction to increase the diversity of positive pairs.


\section{Experiments}

\subsection{Datasets}
Following \citet{DBLP:conf/emnlp/HoganLS23}, we adopt FewRel \citep{DBLP:conf/emnlp/HanZYWYLS18}, TACRED \citep{DBLP:conf/emnlp/ZhangZCAM17}, and Re-TACRED \citep{Stoica2021ReTACREDAS} datasets to train and evaluate our model. To simulate the OpenRE task in real-world scenarios, we assign $|\boldsymbol{\mathcal{C}}_{novel}|=6$ relations as novel relations for each dataset, and the remaining relations are considered as known relations. For each known relation, we allocate half of its instances to the labeled dataset $\boldsymbol{\mathcal{D}}_l$. The unlabeled dataset $\boldsymbol{\mathcal{D}}_u$ consists of the remaining half known instances along with all instances from novel relations.

FewRel dataset includes additional relation hierarchies. To challenge the generalizability of OpenRE models, we assign each instance its top-level relation as the ground-truth label. We identify six single relations without a parent and designate them as novel relations. For TACRED and Re-TACRED datasets, novel relations are randomly selected from all relations. For more details about each dataset's split, see Table \ref{tab:statistics} and Appendix \ref{sec:novel}. 
\begin{table} [h]
	\centering
    \resizebox{0.95\linewidth}{!}{
	\begin{tabular}{ccccc}
		\hline
		\textbf{Dataset} & $|\boldsymbol{\mathcal{C}}_{known}|$ & $|\boldsymbol{\mathcal{D}}_l|$ & $|\boldsymbol{\mathcal{C}}_{u}|$ & $|\boldsymbol{\mathcal{D}}_u|$\\
		\hline
		  FewRel & 35 & 22050 & 41 & 26250\\
            TACRED & 35 & 10074 & 41 & 11692\\
            Re-TACRED & 33 & 15586 & 39 & 18082\\
		\hline
	\end{tabular}}
\vspace{-0.1in}
	\caption{Statistics of labeled and unlabeled datasets.} \label{tab:statistics}
\end{table}

\begin{table*}
        \setlength\tabcolsep{3pt}
	\centering
	\small
	\resizebox{\textwidth}{!}{
	\begin{tabular}{c|c|ccc|ccc|ccc|c}
		\hline
		\multirow{2}{*}{\textbf{Dataset}} & \multirow{2}{*}{\textbf{Method}} & \multirow{2}{*}{\textbf{P}} & \multirow{2}{*}{\textbf{R}} & \multirow{2}{*}{\textbf{F1}} & \multicolumn{3}{c|}{\textbf{B$^3$}}  &  \multicolumn{3}{c|}{\textbf{V-measure}} & \multirow{2}{*}{\textbf{ARI}}\\ 
            & & & & & Prec. & Rec. & F1 & Hom. & Comp. & F1 & \\
	\hline
        \multirow{7}{*}{FewRel} 
        & ORCA & 0.6095 & 0.6328 & 0.6210 & 0.6347 & 0.4823 & 0.5481 & 0.6335 & 
        0.4848 & 0.5492 & 0.4318 \\
        & MatchPrompt$'$ & 0.7575 & 0.6271 & 0.6862 & 0.3031 & 0.8196 & 0.4426 & 
        0.4036 & 0.7599 & 0.5272 & 0.2394 \\
        & TABs$'$ & 0.7296 & 0.6955 & 0.7121 & 0.9193 & 0.7125 & 0.8028 & 0.9088 & 0.7071 & 0.7953 & 0.7746 \\        
        & HiURE$^*$ & 0.4441 & 0.4260 & 0.4349 & \underline{0.9660} & \underline{0.8147} & \underline{0.8838} & \underline{0.9615} & \underline{0.8042} & \underline{0.8758} & \underline{0.8735} \\
        & AugURE$^*$ & 0.5005 & 0.4770 & 0.4884 & \textbf{0.9720} & 0.7914 & 0.8723 & \textbf{0.9647} & 0.7941 & 0.8711 & 0.8568 \\
        & KNoRD & \underline{0.7701} & \underline{0.7775} & \underline{0.7738} & 0.8230 & 0.6587 & 0.7318 & 0.8286 & 0.6519 & 0.7297 & 0.6945 \\
        & \our & \textbf{0.8606} & \textbf{0.8067} & \textbf{0.8328} & 0.9585 & \textbf{0.8426} & \textbf{0.8968} & 0.9490 & \textbf{0.8206} & \textbf{0.8802} & \textbf{0.8817} \\ 
        \hline 
        \hline
        \multirow{7}{*}{TACRED} 
        & ORCA & 0.6845 & 0.7534 & 0.7173 & 0.7501 & 0.4751 & 0.5817 & 0.7381 & 0.4696 & 0.5740 & 0.4622 \\
        & MatchPrompt$'$ & 0.7145 & 0.5989 & 0.6516 & \textbf{0.9357} & 0.6046 & 0.7345 & \textbf{0.9288} & 0.6468 & 0.7626 & 0.7159 \\
        & TABs$'$ & 0.7650 & 0.8175 & 0.7904 & 0.8908 & 0.5462 & 0.6772 & 0.8937 & 0.6214
        & 0.7331 & 0.6647 \\
        & HiURE$^*$ & 0.4976 & 0.4699 & 0.4831 & 0.8908 & 0.7289 & 0.8003 & 0.9010 & 0.7520 & 0.8194 & 0.7953 \\
        & AugURE$^*$ & 0.4989 & 0.4751 & 0.4867 & 0.8966 & \underline{0.7743} & \underline{0.8309} & 0.9071 & \underline{0.7718} & \underline{0.8340} & \underline{0.8001} \\
        & KNoRD & \underline{0.8404} & \underline{0.8638} & \underline{0.8519} & 0.8860 & 0.6778 & 0.7680 & 0.8967 & 0.7033 & 0.7883 & 0.7193 \\
        & \our & \textbf{0.8624} & \textbf{0.9052} & \textbf{0.8833} & \underline{0.8973} & \textbf{0.8429} & \textbf{0.8682} & \underline{0.9081} & \textbf{0.8182} & \textbf{0.8599} & \textbf{0.8473} \\
        \hline 
        \hline
        \multirow{7}{*}{Re-TACRED} 
        & ORCA & 0.6578 & 0.7520 & 0.7018 & 0.6782 & \underline{0.7810} & 0.7260 & 0.6388 & 0.6783 & 0.6579 & 0.5552 \\
        & MatchPrompt$'$ & 0.7160 & 0.5564 & 0.6262 & \underline{0.9875} & 0.5416 & 0.6995 & \underline{0.9805} & 0.6301 & 0.7672 & 0.6223 \\
        & TABs$'$ & 0.5976 & 0.6056 & 0.6015 & 0.9715 & 0.5054 & 0.6649 & 0.9653 & 0.6136 & 0.7503 & 0.5582 \\
        & HiURE$^*$ & 0.4341 & 0.4041 & 0.4185 & 0.9721 & 0.7174 & 0.8253 & 0.9694 & 0.7250 & 0.8294 & 0.8494 \\
        & AugURE$^*$ & 0.4551 & 0.4313 & 0.4429 & \textbf{0.9942} & 0.7575 & \underline{0.8596} & \textbf{0.9908} & \underline{0.7639} & \textbf{0.8625} & \underline{0.8767} \\
        & KNoRD & \underline{0.8493} & \underline{0.8853} & \underline{0.8669} & 0.9698 & 0.4763 & 0.6389 & 0.9583 & 0.5903 & 0.7306 & 0.5081 \\
        & \our & \textbf{0.8972} & \textbf{0.9349} & \textbf{0.9156} & 0.9779 & \textbf{0.7918} & \textbf{0.8750} & 0.9718 & \textbf{0.7733} & \underline{0.8613} & \textbf{0.8925}\\
        \hline
        \end{tabular}}
	\vspace{-0.1in}
 \caption{Performance of all methods on FewRel, TACRED, and Re-TACRED datasets. Precision (P), Recall (R), and F1 score are reported on ground-truth known instances. $B^3$, V-measure, and ARI evaluate the clustering performance on ground-truth novel instances. The details of baseline methods can be found in Sec. \ref{sec:baselines}.}\label{tab:main}
\end{table*}

\subsection{Baselines} \label{sec:baselines}

We compare the proposed model \our with the following state-of-the-art OpenRE methods: (1) ORCA \citep{DBLP:conf/iclr/CaoBL22}, (2) MatchPrompt \citep{DBLP:conf/emnlp/WangZLLZW22}, (3) TABs \citep{DBLP:conf/emnlp/LiJH22}, (4) HiURE \citep{DBLP:conf/naacl/LiuHZLWY22}, (5) AugURE \citep{DBLP:conf/emnlp/WangZQLL23}, and (6) KNoRD \citep{DBLP:conf/emnlp/HoganLS23}. Except for KNoRD, these baselines are not inherently designed for the generalized OpenRE setting and therefore require extensions. The extended baselines and related modifications are discussed as follows.

ORCA: As a computer vision model designed for a similar generalized open-world setting, ORCA does not require structural modifications. It is adapted to the relation extraction task by replacing ResNet with DeBERTa \citep{DBLP:conf/iclr/HeLGC21} and generating relation representations. 

MatchPrompt$'$ and TABs$'$: The OpenRE methods MatchPrompt and TABs are inherently limited in their ability to differentiate between known and novel instances within unlabeled data. To address this, we treat all relations as novel and allow these models to effectively cluster the unlabeled data. We then apply the Hungarian Algorithm \citep{DBLP:books/daglib/p/Kuhn10} to align some clusters with known relations, enabling performance evaluation on both known and novel relations.

HiURE$^*$ and AugURE$^*$: The original HiURE and AugURE models both operate in an unsupervised manner. For fair comparisons, we incorporate a supervised cross-entropy loss in addition to their overall loss function to help fine-tune their relation encoders. Similarly, we leverage the Hungarian Algorithm to assign clusters to known relations. Additionally, we exclude the use of ChatGPT in the AugURE model.

\subsection{Evaluation Metrics}
We evaluate the model performance on the unlabeled dataset $\boldsymbol{\mathcal{D}}_u$. For instances belonging to ground-truth known relations, we measure the performance using precision, recall, and F1 score. For ground-truth novel relation instances, we evaluate clustering performance using $B^3$ \citep{Bagga1998EntityBasedCC}, V-measure \citep{DBLP:conf/emnlp/RosenbergH07}, and Adjusted Rand Index (ARI) \citep{Hubert1985ComparingP}. 
For all of these metrics, higher values indicate better performance.
\begin{itemize}
    \item $B^3$ precision and recall measure the quality and coverage of relation clustering, respectively. $B^3$ F1 score is computed to provide a balanced evaluation of clustering performance.
    \item V-measure is another widely used metric for evaluating clustering quality. Unlike $B^3$, which treats each instance individually, V-measure evaluates both intra-cluster homogeneity and inter-cluster completeness, offering a more comprehensive assessment of clustering performance by considering the overall structure of the clusters.
    \item Adjusted Rand Index (ARI) measures the level of agreement between the clusters produced by the model and the ground truth clusters. It ranges from $[-1,1]$, where a value close to 1 indicates strong agreement, 0 represents random clustering, and negative values suggest disagreement.
\end{itemize}

\begin{table*}
        \setlength\tabcolsep{3pt}
	\centering
	\small
	\resizebox{\textwidth}{!}{
	\begin{tabular}{c|ccc|ccc|ccc|c}
		\hline
		\multirow{2}{*}{\textbf{Method}} & \multirow{2}{*}{\textbf{P}} & \multirow{2}{*}{\textbf{R}} & \multirow{2}{*}{\textbf{F1}} & \multicolumn{3}{c|}{\textbf{B$^3$}}  &  \multicolumn{3}{c|}{\textbf{V-measure}} & \multirow{2}{*}{\textbf{ARI}}\\ 
            & & & & Prec. & Rec. & F1 & Hom. & Comp. & F1 & \\
	\hline
    \our & \textbf{0.8606} & \underline{0.8067} & \textbf{0.8328} & \underline{0.9585} & \textbf{0.8426} & \textbf{0.8968} & \underline{0.9490} & \textbf{0.8206} & \textbf{0.8802} & \textbf{0.8817}\\
    $-$ NRD (\textit{pred\_known}) & 0.7374 & \textbf{0.8440} & 0.7871 & - & - & - & - & - & - & - \\
    $-$ NRD (\textit{pred\_novel}) & - & - & - & \textbf{0.9709} & \underline{0.8065} & \underline{0.8807} & \textbf{0.9612} & \underline{0.8016} & \underline{0.8741} & \underline{0.8651} \\
    $-$ Continual Learning & \underline{0.8484} & 0.8056 & \underline{0.8264} & 0.8573 & 0.7830 & 0.8134 & 0.8648 & 0.7746 & 0.8154 & 0.7516 \\
    
    $-$ Clustering Loss $\mathcal{L}_{e}$ & 0.8440 & 0.8063 & 0.8246 & 0.9373 & 0.7972 & 0.8615 & 0.9256 & 0.7771 & 0.8448 & 0.8382 \\
        \hline
        \end{tabular}}
	\vspace{-0.1in}
 \caption{Ablation study on FewRel dataset.}\label{tab:ablation}
\end{table*}

\subsection{Main Results}

We evaluate \our against state-of-the-art baseline models on the FewRel, TACRED, and Re-TACRED datasets. Additional implementation details are provided in Appendix \ref{sec:implementation}. For all models, the average performance of two random runs is reported. The main results are shown in Table \ref{tab:main}.

On ground-truth known relations, \our consistently outperforms the baseline models across all datasets, achieving the highest precision, recall, and F1 score. Notably, MixORE surpasses the previous best OpenRE model, KNoRD, by 5.90\%, 3.14\%, and 4.87\% in F1 score on FewRel, TACRED, and Re-TACRED, respectively. This highlights the effectiveness of \our in improving the classification performance of known relations.

In novel relation clustering, \our demonstrates competitive performance, consistently ranking among the top-performing models. Although other baselines occasionally achieve higher scores on certain metrics, \our exhibits the strongest overall performance, especially on the TACRED dataset, where it attains the highest $B^3$ F1 score, V-measure F1 score, and ARI. Compared to the second-best model, AugURE$^*$, \our achieves improvements of 3.73\%, 2.59\%, and 4.72\% in these metrics on the TACRED dataset, respectively.

These results suggest that \our effectively captures meaningful relation representations while maintaining a balance between known relation classification and novel relation clustering.

\subsection{Ablation Study}

To evaluate the contribution of different components, we conduct an ablation study by systematically excluding specific components. The results on FewRel dataset are presented in Table \ref{tab:ablation}.

To assess the impact of the novel relation detection (NRD) module, we remove all the weakly-labeled novel instances from the training set (referred to as ``$-$ NRD''). Without NRD, the model cannot distinguish between known and novel relations in the unlabeled data, so we present the results as two separate settings: (1) \textit{pred\_known}, where the model assumes all relations are known and performs classification on the unlabeled data, and (2) \textit{pred\_novel}, where the model treats all relations as novel and performs clustering using K-Means algorithm. Subsequently, setting (1) and setting (2) are evaluated against ground-truth known and novel instances, respectively. The results reveal that excluding NRD leads to a notable -4.57\% drop in the F1 score of known relation classification and a slight decline in novel relation clustering performance. This indicates that the weak labels play an essential role in enhancing the discriminative power on the relation classification task.

We also evaluate the performance of \our without the continual learning paradigm, where the model is initially provided with both labeled data and the weakly-labeled novel instances (referred to as ``$-$ Continual Learning''). As a result, we observe a minor decrease in known relation classification performance and a significant drop (-8.34\%, -6.48\%, and -13.01\% in $B^3$ F1 score, V-measure F1 score, and ARI, respectively) in the clustering performance of novel relations.
These results demonstrate that continual learning allows \our to use previously acquired knowledge to more effectively learn novel relations, making it well-suited for dynamic and evolving tasks. 

To study the advantage of incorporating data distribution, we exclude the clustering exemplar loss function $\mathcal{L}_{e}$ from \our's parameter updates (referred to as ``$-$ Clustering Loss $\mathcal{L}_{e}$''). The results show a small decrease in the classification performance of known relations. For novel relation clustering, we see a performance change of -3.53\%, -3.54\%, and -4.35\% in $B^3$ F1 score, V-measure F1 score, and ARI, respectively. This suggests that considering data distribution is beneficial for both known relation classification and novel relation clustering tasks. 

\subsection{Analysis of Clustering-Derived Weak Labels}
\our leverages novel relation detection (Sec. \ref{sec:nrd}) to generate weak labels for novel relations. A natural concern is the potential propagation of clustering errors to final model performance. To better understand this dependency, we evaluate the quality of the weak labels by measuring the number of novel relations successfully identified and cluster purity. Each dataset contains six novel relations, and the results are summarized in Table \ref{tab:weak_label}. 

\begin{table} [h]
	\centering
    \resizebox{0.95\linewidth}{!}{
	\begin{tabular}{ccc}
		\hline
		\textbf{Dataset} & \textbf{\# Identified Novel Relations} & \textbf{Purity}\\
		\hline
		  FewRel & 5 & 0.608\\
            TACRED & 3 & 0.556\\
            Re-TACRED & 3 & 0.636\\
		\hline
	\end{tabular}}
\vspace{-0.1in}
	\caption{Quality of Clustering-Derived Weak Labels.} \label{tab:weak_label}
\end{table}

Cluster purity is calculated as:
\begin{align}
    \text{Purity} & = \frac{1}{N_o} \sum_{i=1}^{|\boldsymbol{\mathcal{C}}_{novel}|} \max_{j} \left| C_i \cap L_j \right|,
\end{align}
where $N_o$ is the total number of detected outliers, $C_i$ is the set of data points in cluster $i$, and $L_j$ is the set of data points belonging to ground-truth class $j$. 
While the weak labels are not perfectly accurate, they provide useful guidance for modeling novel relations. We observe strong and consistent final performance across all datasets despite moderate purity levels. This suggests MixORE’s robustness to label noise and highlights its ability to learn meaningful representations even under imperfect supervision.

\section{Conclusion}

This paper explores the generalized OpenRE task and introduces MixORE, a two-phase framework that jointly optimizes relation classification and clustering. \our effectively learns discriminative features for known relations while progressively integrating novel information from unlabeled data. Experiments on three benchmark datasets show the superiority of \our over competitive baselines, highlighting its effectiveness in balancing known relation classification and novel relation discovery. 
Our work advances the OpenRE task by introducing a more adaptable approach and offering valuable insights for both future research and real-world applications.

\section*{Limitations}
While our proposed framework demonstrates strong performance in generalized OpenRE, it has certain limitations that call for further exploration.

One limitation of our approach is that it cannot automatically determine the number of novel relations present in the unlabeled data. Instead, it relies on a pre-defined number of clusters, which may not always align with the true distribution of novel relations. Future work could explore adaptive clustering techniques to dynamically estimate the number of novel relations, enhancing the flexibility and applicability of our framework.

Another limitation stems from our implicit assumption that relations are independent of each other. In reality, relations may have hierarchical dependencies, such as being child or parent relations of other relations. Our current method does not explicitly model these dependencies, which may lead to suboptimal performance. Future research could incorporate relational hierarchies into the learning process, enabling a more comprehensive understanding of relation dependencies and improving the model’s ability to handle complex relation structures.

\section*{Ethics Statement}
We comply with the ACL Code of Ethics.

\section*{Acknowledgments}
The work was supported in part by the US National Science Foundation under grant NSF-CAREER 2237831. We also want to thank the anonymous reviewers for their helpful comments.

\bibliographystyle{acl_natbib}
\bibliography{custom}

\newpage
\appendix

\section{Appendix}
\label{sec:appendix}

\subsection{Novel Relations in Each Dataset} \label{sec:novel}

The six single relations without a parent we used as FewRel novel relations are as follows:
\begin{table}[h]
    \centering
    \begin{tabular}{|c|}
        \hline
        ``publisher''  \\ \hline
        ``nominated for''  \\ \hline
        ``instrument''  \\ \hline
        ``notable work'' \\ \hline
        ``competition class'' \\ \hline
        ``position played on team/speciality'' \\ \hline
    \end{tabular}
\end{table}

The randomly selected novel relations from TACRED are as follows:
\begin{table}[h]
    \centering
    \begin{tabular}{|c|}
        \hline
        ``per:city\_of\_birth'' \\ \hline
        ``org:stateorprovince\_of\_headquarters''\\ \hline
        ``org:member\_of''\\ \hline
        ``per:date\_of\_death''\\ \hline
        ``per:city\_of\_death''\\ \hline
        ``per:children''\\ \hline
    \end{tabular}
\end{table}

The randomly selected novel relations from Re-TACRED are as follows:
\begin{table}[h]
    \centering
    \begin{tabular}{|c|}
        \hline
        ``per:siblings''\\ \hline
        ``org:founded\_by''\\ \hline
        ``org:city\_of\_branch''\\ \hline
        ``per:countries\_of\_residence''\\ \hline
        ``per:date\_of\_birth''\\ \hline
        ``per:city\_of\_death''\\ \hline
    \end{tabular}
\end{table}

\subsection{Implementation Details} \label{sec:implementation}

In the first phase, we set the weighting coefficient to $\lambda=100$. During the second phase, we optimize the loss using AdamW \citep{DBLP:conf/iclr/LoshchilovH19}. The encoder is warmed up for $2$ epochs and continually trained for $5$ epochs, all with a learning rate of $1e-5$. We set the margin for the triplet margin loss on labeled data to $\gamma=0.75$. For the clustering exemplar loss function, we use a temperature parameter of $\tau=0.02$ and include $J=10$ negative examples. We implement the granularity layer with $L=4$, setting $c_l \in [16,32,41,64]$ for FewRel and TACRED, and $c_l \in [16,32,39,64]$ for Re-TACRED. All experiments are conducted on an NVIDIA Tesla V100 GPU.

This work and its associated artifacts are licensed under the Creative Commons Attribution 4.0 International (CC BY 4.0) License, allowing unrestricted use, distribution, and reproduction, provided the original work is properly cited using standard academic practices.

\end{document}